# A Probabilistic Analysis of Marker-Passing Techniques for Plan-Recognition

Glenn Carroll
Department of Computer Science
Brown University

Eugene Charniak
Department of Computer Science
Brown University

## Abstract

Useless paths are a chronic problem for marker-passing techniques. We use a probabilistic analysis to justify a method for quickly identifying and rejecting useless paths. Using the same analysis, we identify key conditions and assumptions necessary for marker-passing to perform well.

## 1 Introduction

A recognition problem is one of inferring the presence of some entity from some input, typically from observing the presence of other entities and the relations between them. We will make the common assumption that high-level recognition is accomplished by selecting an appropriate *schema* from a schema library. A schema is a generalized internal description of a class of entities in terms of their parts, their properties, and the relations between them. In the schema selection paradigm, to recognize a "foo" in the input is to create a schema instance foo1 of type foo and assign a high degree of belief in the proposition that foo1 exists. (Henceforth we will assign the degree of belief in the existence of a schema instance to the proposition that the instance is of the appropriate type, e.g., that foo1 is of type foo.) In plan recognition, the generalized plans are schemas. While the system which we will discuss has been applied to plan recognition in the context of story understanding, we will continue to talk of schema, since we wish to emphasize that our system is applicable to high-level recognition in general.

A crucial problem faced by schema selection is that of searching the schema library for the right schema; typically a single piece of local evidence is multiply ambiguous as to the schema which it could indicate. For example, an act of getting a rope might fit into many schemas.

One of the few concrete suggestions here has been *marker passing* (Alterman [1985]; Charniak [1983]; Charniak [1986]; Collins & Quillian [1969]; Hendler [1988]; Norvig [1987a]). Marker-passing uses a breadth-first search to find paths between concepts in an associative network made up of concepts and their part-subpart relations. In our case, the concepts will be schemas, i.e. plans and/or actions. The idea is that a path between two schemas suggests which schema(s) to consider for recognition. For example, a knob instance and a hinge instance might suggest a door (instance); since there are links between the schemas door and knob and between door and hinge in the associative network (they are part-subpart relations), there is therefore a path from knob to door to hinge. Unfortunately, most marker-passer systems have found many more bad paths, suggesting incorrect schemas, than good ones (Charniak [1986]; Norvig [1987b]). We will show in this paper that the good/bad path ratio can be raised quite high by exploiting probability information; we realize this benefit by (cheaply) controlling the marker-passer's search, both extending it in promising directions and terminating it in unpromising ones.

In this paper we will give a probabilistic account of marker passing. This account will have two goals — first it should shed further light on when marker-passing is an appropriate technique, and second it should show how to improve the performance of marker-passing algorithms by increasing the likelihood that the paths generated will, in fact, suggest the correct schema. In section 2 we will consider schema evaluation within a probabilistic framework. That is, given we have a potential schema, how do we evaluate the probability that it is the correct explanation of the input. In particular, we will be adopting a Bayesian network (or belief network) formulation of the problem, so the probability distributions correspond to DAGs with probabilities associated with each node. Section 3 will then be concerned with schema selection, i.e., how our marker passing system works, and how the schema suggestions (paths) from the marker-passer map to Bayesian networks. In Section 4 we will show how to use probability information from the knowledge base to intelligently limit the marker-passing search. Specifically, we will describe how to calculate on the fly a measure which is an upper bound on the joint probability of the schemas which a candidate path suggests. Thanks to properties of the marker-passer paths and our probabilistic model, we can avoid constructing and evaluating a Bayesian network to evaluate each path, an NP-hard problem (Cooper [1987]), so our evaluation need not be expensive. This section has the bulk of the new work in the paper. We summarize and explain results in section 5.

The opening sections of this paper alternate somewhat irregularly between the marker-passer and the Bayesian network; while it might appear to be simpler to fully describe first one and then the other, this would leave much of what we have to say completely unmotivated and very likely obscure. Once we reach section 4, we treat the two systems together, showing how paths map to Bayesian networks, and how path calculations yield an upper bound on the joint probability of the nodes in the Bayesian network.

## 2  Probabilistic Schema Evaluation

We adopt a standard first-order theory of schema in which a schema is a set and asserting that an entity is an instance of that schema is asserting that it is an element of the set. We use the predicate inst for this purpose.

(inst *instance schema*).

Schemas are related in the usual isa-hierarchy (subset) as in

(isa *specific-schema general-schema*).

In this paper we assume that isa relations form a tree, not a lattice, and thus all the immediate isa subsets of a given parent are disjoint.

Slots or roles in schemas are represented using functions from a schema instance to the slot-filler schema instance. Equality is used to assert that a particular entity fills that role. For example, to assert that a particular store store-25 fills the store-of role in supermarket-shopping-3 we assert

(== (store-of supermarket-shopping-3) store-25)

Facts about the relations between the parts of a schema are then universally quantified facts about the corresponding functions. For example, to say that every store-of a supermarket-shopping must be filled with an instance of a supermarket (another schema) we would say

(inst ?x supermarket-shopping) → (inst (store-of ?x) supermarket)

To abbreviate this we will write: (role supermarket-shopping store-of supermarket). More generally,

(role $schema_1$ $slot$ $schema_2$)

states that anything which fills *slot* in $schema_1$ must be an instance of $schema_2$. Note that role is not a predicate of our plan recognition language, but is rather an abbreviation for formulas of the above form.

In our probabilistic version we will determine the probability of a plan by embedding inst and == statements in a Bayesian network. We will not attempt to summarize Bayesian networks but rather will assume the reader has a working knowledge of them. (See (Pearl [1988]) for a good introduction.) Equality (==) statements will become random variables with possible values 1 and 0 (true and false). inst statements become random variables which can take any maximally specific schema type as their value[1]. Thus the probability of the statement (inst sms1 supermarket-shopping) would become the probability that the instance sms1 takes on the value supermarket-shopping. However, often we will talk as if the statement (inst sms1 supermarket-shopping) appears in the network (with values 1 and 0). Most of the time the two representations are interchangeable.

We intend our prior and conditional probabilities to come from a sample space of explanations for some large corpus of stories. For example, the prior probability of a supermarket-shopping plan would be the number of supermarket-shopping plans that appear in our set of explanatory plans, divided by the total number of explanatory plans. See (Goldman [1991]) for a detailed description of the probability model.

## 3  Probabilistic Schema Selection

### 3.1  Marker-Passing

Marker-passing searches for paths between schemas in a graph whose nodes are schemas and whose arcs are isa and role statements. Marker-passing works as follows: the marker-passer is given some schema, derived from a new inst statement, e.g., (inst supermarket1 supermarket). It places a mark on that schema, and then proceeds in breadth-first order to place marks on all the neighbors in the graph, their neighbors' neighbors, and so on. For example, our supermarket schema has two neighbors, supermarket-shopping, which is connected by the statement (role supermarket-shopping store-of supermarket) and store-, which is connected by the statement (isa supermarket store-). Both of these would be marked after supermarket, and then their neighbors would be marked. Each mark has a numeric value, which generally diminishes according to its distance from the original mark (c.f., "zorch" in (Charniak [1986])). This value is used to cut off marker-passing, since otherwise we would continue until the entire graph was covered. For our value, we use an upper bound on the schemas and relations suggested by the path; we will be precise in section 4. After marking a node, the marker-passer checks for marks from some other origin on the same node. If such a mark is found, both it and the new mark are back-traced to their respective origins, and the resulting lists of statements are glued together to form a path. For example, suppose we found a mark on supermarket-shopping which had originated at go. We would have as a path:

```
(inst supermarket2 supermarket)
(role supermarket-shopping store-of
 supermarket)
(isa shopping supermarket-shopping)
(role shopping go-step go)
(inst go1 go)
```

---

[1]Thus, the sum over all possible values is 1.0, which would not be true if they could take on non-maximally specific types.

We include the original inst statements, even though the marker-passer does not, strictly speaking, pass marks over these links. It will be convenient for us to refer to them as part of the path, and they serve to disambiguate this path from other paths which may have the same links but different origins.

The marker-passer returns a list of all the paths which it found once the marking has terminated. For more detail on how marker-passing works, see (Hendler [1988]).

### 3.2 Valid Paths and Interpretations

Intuitively, we wish to interpret a path as a claim about how its ends are related to each other. In order to do this, we need to translate the path through the semantic network (a list of inst, isa, and role statements) into a set of Bayesian network nodes (inst and equality statements) and arcs. Before we describe this mapping, we must confess that our marker-passing system is not precisely the very simple one described above. We employ a DFA at each node in our network to control the marker-passing, allowing us to restrict the form of paths which we generate and report. This allows us to skip paths which are malformed in the sense that either they cannot be translated into a consistent set of statements in our schema theory, or they embody demonstrably bad schema suggestions. A valid path is one which is not malformed in the above sense.

NOTATION. By isa- we mean isa with the order of arguments reversed. So,

(isa *specific-frame general-frame*) iff (isa- *general-frame specific-frame*)

Similarly for role and role-:

(role *filler-frame filled-frame slot*) iff (role- *filled-frame filler-frame slot*)

In future discussions we will often fail to distinguish between the predicates and their "-" versions.

DEFINITION 3.1. *A valid path from $i_1$ to $i_2$ has the form*

(inst $i_1$ $s_1$) (pred$_1$ $s_1$ $s_2$) ... (pred$_n$ $s_n$ $s_{n+1}$)
(inst $i_2$ $s_{n+1}$)

*where*

- *pred$_i$ may be one of* isa, isa-, role *and* role-,
- *at least one* role *appears among the pred$_i$.*
- *no sequence* (isa ...)(isa- ...) *appears among the pred$_i$*
- *if pred$_i$ is a* role-, *then no pred$_k$ where $k > i$ can be a* role

Our last two restrictions prohibit *isa-plateaus*, where an isa is followed by an isa-, and slot-filler valleys, where a role- is followed by a role, possibly with isa's between them. We have calculated, off-line, the joint probability of the statements associated with paths which violate the above restrictions; in all cases the joint probability falls below our threshold for being worth computing[2].

We will now define the statements associated with a path $P$, written $S(P)$.

NOTATION. By $P[n]$ we mean the nth statement of a path $P$.

The *relevant instance* at $P[n]$ (written $I(P[n])$) is defined as follows:

DEFINITION 3.2.

1. *If (inst i s) = $P[n]$, then $I(P[n]) = i$.*
2. *If (isa $s_1$ $s_2$) = $P[n]$ and $I(P[n-1]) = i$, then $I(P[n]) = i$. Similarly for isa-.*
3. *If (role $s_1$ slot $s_2$) = $P[n]$ then $I(P[n]) = i'$, where $i'$ is a new constant term.*

$S(P)$ is defined as follows:

DEFINITION 3.3.

1. *If (inst i t) = $P[n]$, then (inst i t) $\in S(P)$.*
2. *If (isa $s_1$ $s_2$) = $P[n]$ and (inst i t) $\in S(P)$, then (inst i $s_2$) $\in S(P)$. Similarly for isa-.*
3. *If (role $s_1$ slot $s_2$) = $P[n]$, then $\{$(inst $i_n$ $s_2$) (==(slot $i_n$) i))$\} \subset S(P)$, where $i_n = I(P[n])$. Similarly for role-.*

Intuitively we wish to interpret a path $P$ as a claim about how its ends are related to each other. "Every $s \in S(P)$ is true" is intended to be the formalization of

---

[2]To summarize: for isa-plateaus, the left and right side of the Bayesian net in which we embed the statements are independent, so they cannot support one another. For the slot-filler valleys, there is no evidence to support the object which must appear in two schemas. See (Charniak & Carroll [1991]) for details.

```
(inst supermarket2 supermarket)
(role supermarket-shopping store-of
supermarket)
(isa- shopping supermarket-shopping)
(role go shopping go-step) \nopagebreak
(inst go1 go)
```

FIG. 3.1. An Example Marker-passer Path.

```
(inst supermarket2 supermarket)
(== (store-of shopping3) supermarket2)
(inst shopping3 supermarket-shopping)
(inst shopping3 shopping)
(== (go-step shopping3) go1)
(inst go1 go)
```

FIG. 3.2. $S(P)$ For Example Path.

this claim. For example, the path in Figure 3.1 would have as its $S(P)$ the statements shown in Figure 3.2. We actually need only a subset of $S(P)$, namely the *relevant statements* associated with $P$, (written $RS(P)$) which we will define below. First we give two more definitions necessary for defining $RS(P)$.

We define isa* as

DEFINITION 3.4. $\forall t, t'$ *(isa\* t t') iff ((isa t t') or $\exists t''$ [(isa t t'') and (isa\* t'' t')]).*

We define the *relevant type* of an inst (written $RT(i)$) to be the most specific schema type of the node; more formally,

DEFINITION 3.5. $RT(i) = t$ *such that* (inst $i$ $t$) $\in S(P)$ *and* $\forall t'$ ((inst $i$ $t'$) $\in S(P) \rightarrow$ (isa\* t t') or $t = t'$).

We define the *relevant statements* associated with $P$ as

DEFINITION 3.6.

1. *If (inst i t)* $\in S(P)$ *and* $RT(i) = t$ *then (inst i t)* $\in RS(P)$.
2. *If (== (slot i) j)* $\in S(P)$, *then (== (slot i) j)* $\in RS(P)$.

In effect, we remove superfluous instance statements from $S(P)$, whose statements are still implied by those retained in $RS(P)$.

Our formal measure of a path $P$ is defined by embedding the members of $RS(P)$ in a Bayesian network, and then evaluating the probability that each node is true, given the evidence. In general there may be several paths for the same entities, indicating alternative possible plans. In what follows we will be looking at Bayesian networks in which there is only one $RS(P)$. The idea is that we are interested in getting a preliminary guess as to how likely a particular interpretation— an $RS(P)$– is, and we can do this without detailed comparisons with its competitors.

### 3.3 Vertebrate Bayesian Networks

With each path $P$ we will associate a Bayesian network with a particular structure which will prove important to our calculations. We call such networks *vertebrate* Bayesian networks, because they have *spines*.

DEFINITION 3.7. *A Bayesian network is a "vertebrate" Bayesian network iff it consists of two parts, to be defined below, called the* spine, *and the* interior.

Intuitively, a spine is the geometrical backbone of its Bayesian network.

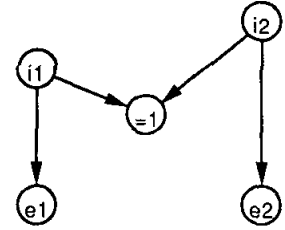

FIG. 3.3. The basic spine.

NOTATION. We use $i_j$ to name inst nodes.

Recall that equality nodes represent slot-filler relationships for us.

NOTATION. We use $=_j$ to name equality nodes, indicating in this case that the parent node $i_j$ is the slot filler.

There are other types of nodes which act as evidence for our equality and inst nodes. For example, the appearance of a word in text provides evidence for the existence of a particular inst. Although these nodes come in several flavors, we can treat them generically for our purposes, so we will name all of them simply "evidence" nodes.

NOTATION. We use $e_j$ to name evidence nodes.

Our definition describes a structure topologically; thus, we imply that a node subscripted by $j$ is not equal to any node subscripted by $k$, where $k \neq j$.

Legal spines are recursively defined as follows:

DEFINITION 3.8.

**Base step:** *Any Bayesian network whose node set is* $\{i_1, i_2, =_1, e_1, e_2\}$ *and whose edge set is* $\{i_1 \rightarrow e_1, i_2 \rightarrow e_2, i_1 \rightarrow =_1, i_2 \rightarrow =_1\}$ *is a spine. See Figure 3.3.*

**Recursion:** *If $S$ is a spine with the nodes $i_1$ and $e_1$ and the edge $i_1 \rightarrow e_1$, then $S'$ is a spine, where*

$$N(S') = N(S) \cup \{i', =_{i'}\} \qquad (1)$$

*and*

$$E(S') = \{E(S) - \{i_1 \to e_1\}\} \cup \\ \{i_1 \to i', \ i' \to e_1, \\ i_1 \to =_{i'}, \ i' \to =_{i'}\} \quad (2)$$

*See Figure 3.4.*

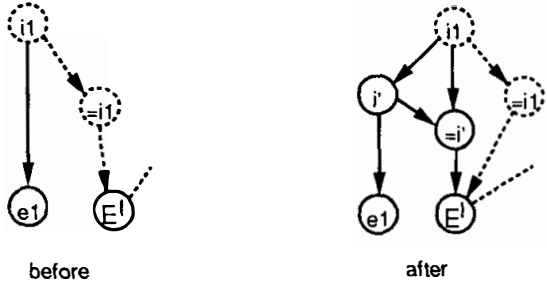

FIG. 3.4. Before and after nodes are added to a spine.

As for the "interior" of a vertebrate Bayesian network, intuitively it is the evidence supporting the equality nodes of the vertebrate Bayesian network. We may assume, without loss of generality that there is only once such evidence node $E^I$. More formally:

DEFINITION 3.9. *If $V$ is a vertebrate Bayesian network with spine $S$, and $S$ has the non-evidence nodes $N$ and the evidence nodes $E$, then the interior of $V$ is an evidence node $E^I$ disjoint from $E$ and a set of edges $D$ from every equality node in $N$ to $E^I$.*

The assumption that there will be supporting evidence, some $E^I$ node, is the crucial one for marker-passing. When there is no evidence, the posterior probabilities of the abductive hypotheses generated from our paths turn out to be abysmally low; plainly put, they are bad guesses. In general, we believe that

CLAIM 3.1. *A domain is suitable for search by means of marker-passing only if there will usually be supporting evidence for paths returned by the marker-passer.*

We will support our claim by showing that, in our domain, which meets the evidence condition, we can increase the ratio of good to bad paths returned from the marker-passer to better than 90%. Our claim should not be interpreted as saying that the marker-passer has no responsibility for the quality of paths it returns; quite the opposite is true. Most of the remainder of this paper will focus on the calculations which allow us to determine whether or not a path is worthwhile, *assuming that there is evidence for it*. If we could not make these calculations, then doubtless many of the paths which would be returned would in fact *fail* to have associated evidence. We can safely throw them out because they are bad paths *regardless* of whether they have evidence.

### 3.4 Relating Paths to Networks

We will now show that each path corresponds to a unique vertebrate Bayesian network, and that the joint probability of $S(P)$ in the network can be calculated from each step of the path.

THEOREM 3.1. *If $P$ is a valid path, then there exists a unique vertebrate Bayesian network $V$ such that the statements in $RS(P)$ have formulas in one-to-one correspondence with the non-evidence nodes of $V$.*

*Proof.* The proof is by induction on the length of the path.

The basis follows from the definition of the simplest spine, and $RS(P)$ for the shortest valid path. The spine and $RS(P)$ have one $==$ node (statement, respectively) and two inst nodes (statements, respectively). For the induction step, suppose we have proved there is a unique network for $P$. Let $P'$ have the same structure as $P$, with an extra isa statement inserted in some location which does not violate our constraints for path validity. We first note that isa statements do not add statements to $RS(P)$, they only change the schemas named in statements already there. Hence, if we have a vertebrate Bayesian network for $P$, we can use the same structure (with different relevant types for some nodes) for $P'$.

If $P'$ is $P$ with an extra role statement (again, inserted in some location which respects path validity), this corresponds to applying clause 2, the recursion step, of the definition of a spine. The role statement adds an $==$ statement and an inst statement to $RS(P)$, and applying clause 2 adds the corresponding nodes to the network. Since each step determines a unique transformation and each transformation results in a unique vertebrate Bayesian network, $P$ corresponds to a unique vertebrate Bayesian network, which we call $V(P)$, or, where unambiguous, just $V$. □

From the above, it should be obvious that we can construct a vertebrate Bayesian network by sequentially processing a path from left to right, adding new nodes and arcs for each role statement we come across, and changing the relevant type of our last added node when we encounter an isa-. We will use this fact, together with some distribution properties of vertebrate Bayesian networks to calculate our measure of path utility without vertebrate Bayesian network corresponding to the path.

## 4 Path Calculations

### 4.1 The Spinal Contribution

Our calculations will not compute the exact joint probability of the network which we construct. Instead we will compute an upper bound on the joint probability, under assumptions to be detailed below, which we call the *spinal contribution*. We will define the spinal contribution momentarily, in terms of the joint probability of the network. To begin with, the exact joint probability of the simplest vertebrate Bayesian network (the basic spine with an added interior evidence node, $E^I$, pictured in Figure 4.1) is

$$\frac{p(e_1|i_1)p(E^I|=_1)p(e_3|i_2)p(i_1)p(i_2)p(=_1|i_1,i_2)}{p(e_1,E^I,e_3)} \quad (3)$$

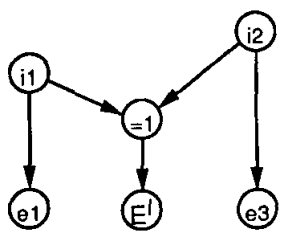

FIG. 4.1. The basic vertebrate Bayesian network.

We use conditional probability and independence to transform the denominator into

$$p(e_1)p(e_3)p(E^I|e_1,e_3) \quad (4)$$

For the numerator, we note that

$$p(e_1|i_1) = p(i_1|e_1)p(e_1)/p(i_1) \quad (5)$$

and similarly for $p(e_3|i_2)$.

The slot-filler term, $p(=_1|i_1,i_2)$, requires some discussion. Recall that role statements specify a particular type for each slot that a schema has. The probability that $i_1$ fills this particular slot in $i_2$ is the probability that any two things of the specified type are equal. This, in turn, is equal to the prior probability of any two things being equal, divided by the prior probability of a thing being the specified type.

NOTATION. We write $p(==)$ for the prior probability of any two things being equal.

This allows us to rewrite the last term in our formula as follows:

$$p(=_1|i_1,i_2) = p(==)/p(i_1) \quad (6)$$

Applying the substitutions in equations 4, 5, and 6 to 3, and then cancelling and regrouping yields

$$\left(\frac{p(i_1|e_1)p(i_2|e_3)}{p(i_1)}\right)\left(\frac{p(==)p(E^I|=_1)}{p(E^I|e_1,e_3)}\right) \quad (7)$$

The right-hand group of terms will appear in the exact calculations for every vertebrate Bayesian network, with the difference that $E^I$ may be conditioned on more nodes, if they are present. By our earlier assumptions about the distributions for $E^I$, this group has an upper bound of 1.0.

The left-hand group of terms we call the *spinal contribution* of our joint probability; more generally, any terms not included in the bounded group, will be part of the spinal contribution, and calculating it will be the focus of the rest of this section. Since those terms not in the spinal contribution have an upper bound of 1.0, the spinal contribution is an upper bound on the joint probability of the network.

### 4.2 Calculations

THEOREM 4.1. *As a path P is traversed, our measure of the spinal contribution of the corresponding vertebrate Bayesian network fragment, $SC(V)$, can be computed recursively in the following manner:*

1. *The initial value, corresponding to the (inst $i_1$ $s_1$) node/statement[3] is $p(i_1|e_1)$, our current belief in the node.*
2. *As each subsequent statement is traversed, we compute the new value by multiplying the current value by the number given in table 4.1.*

TABLE 4.1. Spinal Contribution Multipliers

| Link | Multiplier |
|---|---|
| (role $s_1$ slot $s_2$) | $p(s_1)/p(s_2)$ |
| (role- $s_1$ slot $s_2$) | 1.0 |
| (isa $s_1$ $s_2$ ) | 1.0 |
| (isa- $s_1$ $s_2$ ) | $p(s_2)/p(s_1)$ |
| (inst $i_1$ $s_1$ ) | $p(i_1|e_1)/p(s_1)$ |

*Proof.* Omitted due to space limitations. See (Charniak & Carroll [1991]). □

### 4.3 Internal Calculations

Marker-passing produces whole paths as output; internally, however, it builds these from two half-paths which resulted from passing marks from two differing origins (at different times). We would like to use our measure of spinal contribution to cut off the depth of marker-passing, which requires that we compute it as the half-path is built, before the two halves are put together. We now show that this is possible, and that the calculations for an entire path, above, comprise the bulk of the work for computing half-paths. Our lemma concerns the spine, not the entire vertebrate Bayesian network, since the interior evidence node is not part of our spinal contribution.

DEFINITION 4.1. *By* cleaving *a vertebrate Bayesian network graph at some* inst *node n, we mean that*

- *the cleaved node n appears in both halves*

---
[3]This is a node in the Bayesian network, and a statement in the path. Since we will be discussing probabilities from now on, we will generally call them nodes.

- *the left half includes the evidence node $e_1$, all inst nodes $i_1$ through $i_n$ and all arcs between them, and similarly for the right half*
- *equality nodes with both parents among inst nodes $i_1$ through $i_n$ and all arcs incident to them are in the left half, and similarly for the right*

See Figure 4.2.

LEMMA 4.2. *A spine can be cleaved into two halves $H_1$ and $H_2$ at any inst node, such that the spinal contribution of the whole graph is given by*

$$SC(V) = SC(H_1) \times SC(H_2)/p(n) \qquad (8)$$

*where $n$ is the node at which the graph is cleaved.*

*Proof.* Omitted due to space limitations. See (Charniak & Carroll [1991]). □

Lemma 4.2 means that we can track the spinal contribution incrementally as we extend a half-path. When the measure drops below a threshold, $T$, we can cut safely cut off marker-passing; that is, we will miss no complete paths whose measure would be above $T^2$. Currently we have $T$ set at 30. We have arrived at this value through experimentation with a set of paths generated from a set of stories which we use for debugging and tuning our system. Generally, there is a large gap, a factor of 10 or more, between the spinal contribution of those paths which have the right explanations and those which do not. While our system does rely on the prior probabilities for our schema, this gap suggests that we can get by with priors that are only approximately correct.

## 5 Results

We have employed our marker-passer in the Wimp3 story understanding system (Goldman [1991]) to find explanatory plans. The results quoted in table 5.1 are those obtained both on Wimp3's debugging corpus of 25 1-to-4 line "stories" and on its evaluation corpus[4] of 25 stories. We counted paths at four points in the flow of control. First, we counted paths which left the marker passer. As described above, we integrated the probability valuation of paths into the marker-passing mechanism itself, using it to control the spread of marks. This makes it impossible to estimate how many paths were eliminated due to low probability values. Likewise, we cannot say how many paths were eliminated by employing DFA's to prohibit generation of invalid paths. We can only give the total number of paths returned, with the invalid and low probability paths already weeded out. Second, we counted paths which were "asserted", i.e., used for forward-chaining and Bayesian network construction. These are paths which passed various secondary filters reported in (Carroll & Charniak [1989]). Third, we counted paths whose resulting statements were actually evaluated using our Bayesian network evaluation mechanism. Some paths could be eliminated without evaluation, as we will describe shortly. Finally, we counted those paths which we approved after evaluation; for a path to be approved the posterior probability of the suggested plans given the evidence had to be 1,000 times higher than the prior probability for the plans. The ratio of approved paths to asserted paths was 68% for the debugging corpus, and 59% for the test corpus, a significant improvement on the 10% good to bad path ratio reported earlier, (Charniak Neat 1986 Norvig 1987 Berkeley ]) and strong support for our claim that the key to the viability of marker-passing is the supporting evidence, our $E^I$.

TABLE 5.1. Results Summary

| corpus | paths reported | paths asserted | paths evaluated | paths approved |
|---|---|---|---|---|
| debug | 985 | 115 | 83 | 78 |
| test | 747 | 109 | 68 | 64 |

Our analysis suggested one more improvement which could be made. As we commented earlier, the marker passer's probability calculations are an upper bound, based on the assumption that there is evidence in favor of the path (other than the nodes at either end). That is, after the marker passer produces an acceptable path $P$, Wimp3 constructs a Bayesian network which includes $RS(P)$. Among the paths which were not approved, we found that there was often no evidence supporting some of the statements in $RS(P)$. While we cannot determine whether evidence is missing before network construction, we can do so before network evaluation. While the former takes time linear in the size of the network (making reasonable assumptions about the process), the latter, in general, takes exponential time (and is NP-hard). Indeed, network evaluation accounts for approximately 90% of Wimp3's running time, and thus the only really bad paths are those which cannot be removed before network evaluation and are not approved afterwards. Weeding out paths with no evidence accounts for the difference between paths which were asserted–used for network construction–and paths which were evaluated. A combined total of 151 paths had to be evaluated by Wimp3, and, of these, all but 9 were good, for a percentage of about 94%. This leads us to extend our claim to say that: *A domain is suitable for search by means of marker-passing only if there will usually be supporting evidence for paths returned by the marker-passer, or paths without evidence can be cheaply identified.*

---

[4]The debugging corpus is used for testing and tuning of parameters. The evaluation corpus is for evaluation only, and is therefore a cleaner test, in some sense.

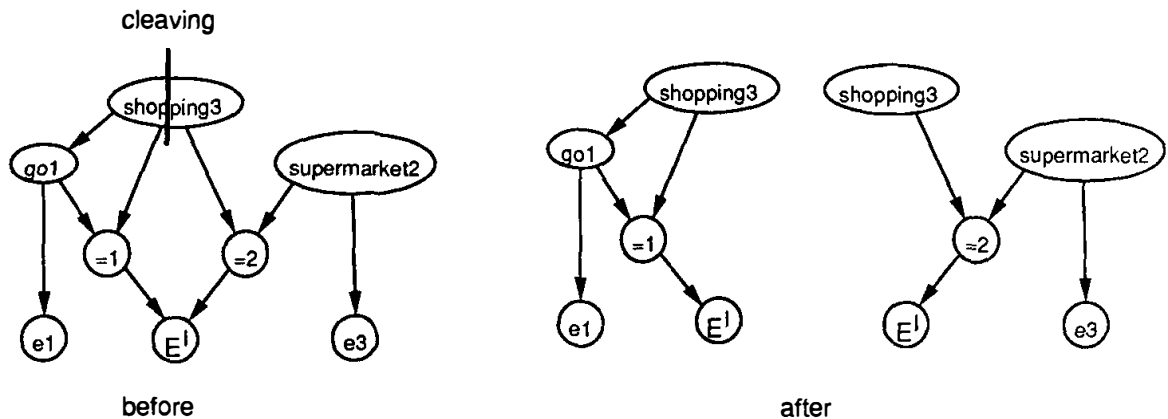

FIG. 4.2. Cleaving a vertebrate Bayesian network.

Acknowledgements

This work has been supported by the National Science Foundation under grant IRI-8911122 and by the Office of Naval Research, under contract N00014-88-K-0589.